\documentclass[11pt,a4paper]{article}
\usepackage[hyperref]{eacl2021}
\usepackage{times}
\usepackage{latexsym}

\usepackage{microtype}

\usepackage{amssymb} 
\usepackage{booktabs}
\usepackage{graphicx}
\usepackage[symbol]{footmisc}

\usepackage{xcolor,colortbl}
\aclfinalcopy 

\title{On the Effectiveness of Dataset Embeddings in Mono-lingual, Multi-lingual and Zero-shot Conditions}

\author{Rob van der Goot\Thanks{$\;$ Equal contributions} \\
  IT University of Copenhagen \\
  \texttt{robv@itu.dk} \\ \And
  Ahmet \"{U}st\"{u}n\footnotemark[1]\\
  University of Groningen \\
  \texttt{a.ustun@rug.nl} \\ \And
  Barbara Plank \\
  IT University of Copenhagen \\
  \texttt{bapl@itu.dk}}

\date{}

\begin{document}
\maketitle
\renewcommand{\thefootnote}{\arabic{footnote}}
\begin{abstract} 
Recent complementary strands of research have shown that leveraging information
on the data source through encoding their properties into embeddings can lead
to performance increase when training a single model on heterogeneous data
sources.  However, it remains unclear in which situations these \textit{dataset
embeddings} are most effective, because they are used in a large variety of
settings, languages and tasks.  Furthermore, it is usually assumed that gold
information on the data source is available, and that the test data is from a
distribution seen during training.  In this work, we compare the effect of
dataset embeddings in mono-lingual settings, multi-lingual settings, and with
predicted data source label in a zero-shot setting.  We evaluate on three
morphosyntactic tasks:\ morphological tagging, lemmatization, and dependency
parsing, and use 104 datasets, 66 languages, and two different dataset grouping
strategies.  Performance increases are highest when the datasets are of the
same language, and we know from which distribution the test-instance is drawn.
In contrast, for setups where the data is from an unseen distribution,
performance increase vanishes.\footnote{source code is available at: 
\url{https://bitbucket.org/robvanderg/dataembs/src}}
\end{abstract}

\section{Introduction}
The performance of natural language processing systems is dependent on the
amount of training data, which is often scarce. To complement existing training
data, supplementary data sources can be used. Especially data annotated for the
same task from other sources can be beneficial to exploit. However, because of
heterogeneity in language or domain this might lead to sub-optimal performance.
In early work on combining training sources, data was selected at training
time~\cite{plank-van-noord-2011-effective,khan-etal-2013-towards} for a given
test set.  A more nuanced way to exploit heterogeneous data is to encode
properties of the language as features~\cite{naseem-etal-2012-selective}.

Recently, ~\newcite{ammar-etal-2016-many} showed that encoding the language of
an instance as an embedding in a neural model is beneficial for multi-lingual
learning.\footnote{More recently,~\cite{conneau2019cross} showed that embedding
the language can also be beneficial for training contextualized embeddings with
masked language modeling.} Follow-up work found that multiple datasets within
the same language can also be combined by encoding their
origin~\cite{stymne-etal-2018-parser,ustun-etal-2019-multi}, thereby implicitly
learning useful commonalities, while still encoding dataset-specific knowledge.
These \textit{dataset embeddings} are employed in groups of datasets which
usually range in size from 2 to 10 datasets.  However, it remains unclear in
which situations these dataset embeddings thrive best.

Furthermore, two often overseen issues with dataset embeddings are that they
are commonly learned from the \textit{gold} data-source labels attached to each
training and test instance and it is assumed that the test data is from a
distribution which is seen during training. In many real world situations these
assumptions are clearly violated.  A common strategy when the test data is
drawn from a different distribution as the training datasets (zero-shot), is to
use a manually assigned proxy
treebank~\cite{smith-etal-2018-82,barry-etal-2019-cross,meechan-maddon-nivre-2019-parse}.
Recent work showed that for unseen datasets in mono-lingual
setups~\cite{wagner-etal-2020-treebank}, interpolated dataset embeddings can be
used to improve performance for zero-shot settings. We use automatically
predicted proxy data sources instead, and focus on mono-linugal as well as
cross-lingual setups.

In this paper, we provide an extensive evaluation of the usefulness of dataset
embeddings in existing setups and beyond. More concretely, we ask: 1) What are
good indicators to predict the usefulness of dataset embeddings? 2) Can we
effectively use dataset embeddings in the absence of gold data-source
information?

\section{Dataset Embeddings}

\begin{figure}
\centering
  \includegraphics[trim=75 20 75 0,clip,width=.95\linewidth]{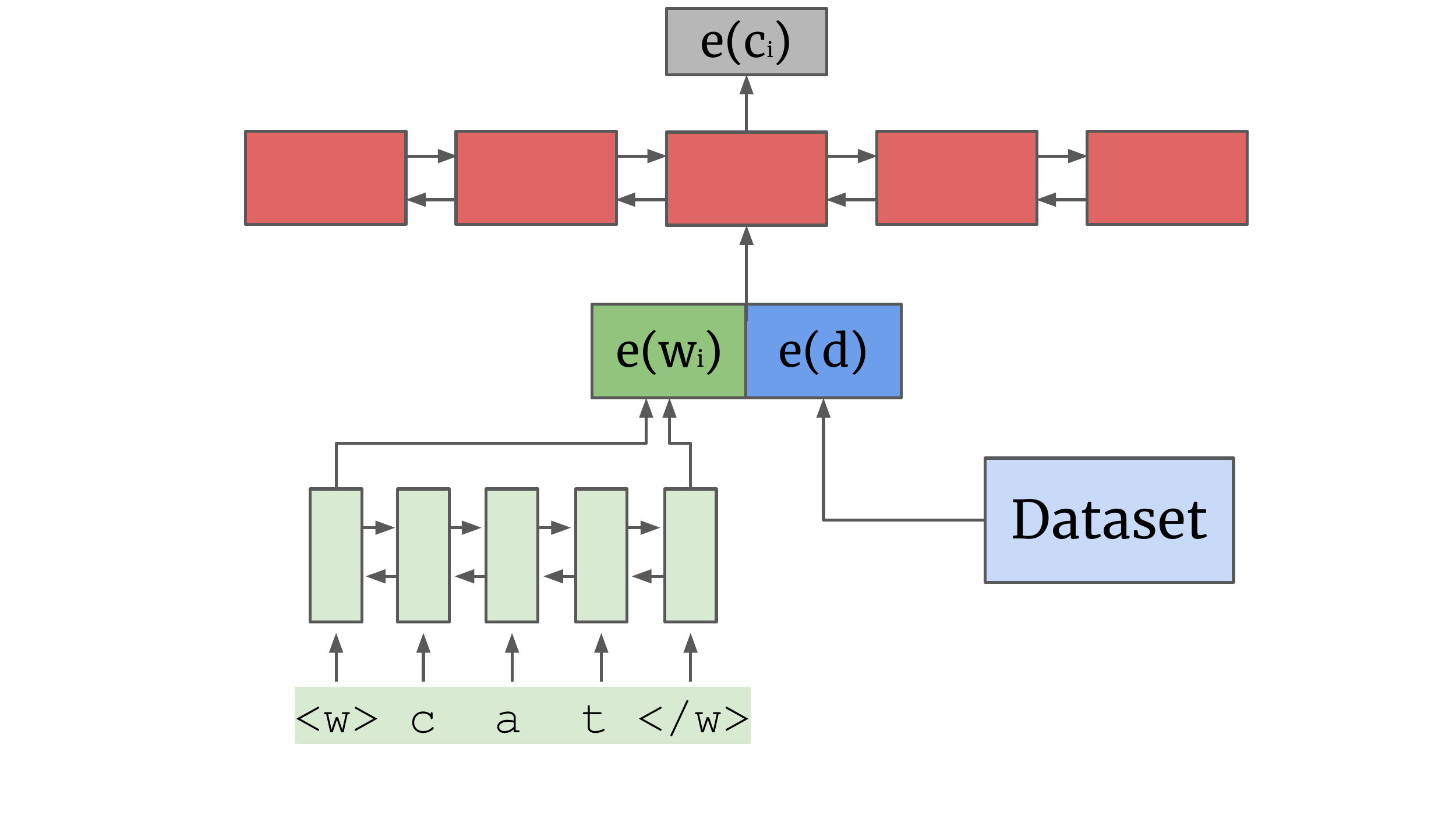}
  \caption{Overview of the model where a BiLSTM encodes the word ``cat" in a sentence. Dataset embeddings (blue) are concatenated with the character-based word representation (green) which feeds into the contextual encoder (red).}
  \label{fig:encoder}
\end{figure}

Dataset embeddings enable conditioning of inputs on some property of the data
when training on multiple sources. They are vector representations learned
during model training, with the aim to capture distinctive properties of the
sources into a continuous vector,  without losing their heterogeneous
characteristics. Given $D$ data sources, technically, we learn a vector
representation $e(d)$ for each data source $d \in D$ while training a single
model from a group of sources. Every input instance 
marked with its dataset source $d$.  For each word $w_i$ with $i=0, ..., n$,
the word embedding $e(w_i)$ is concatenated with the dataset embedding $e(d)$,
and both are updated during training.  Figure 1 shows the overall architecture
of the model which employs a contextual encoder that uses the resulting
embedding as input and outputs $e(c_i)$ to be used for prediction.

\subsection{Experimental Setup}
In this work, we copy the exact setups from the UUParser
2.3~\cite{smith-etal-2018-82} and the Multi-Team
tagger~\cite{ustun-etal-2019-multi} because they were high ranking systems in
two recent shared
tasks~\cite{zeman-etal-2018-conll,mccarthy-etal-2019-sigmorphon} and they both
showed large gains by using dataset embeddings.  The UUParser is an Arc-Hybrid
~\cite{kuhlmann-etal-2011-dynamic} BiLSTM~\cite{graves2005framewise} dependency
parser, which exploits a dynamic oracle~\cite{goldberg2013training} and
supports non-projective parsing through the use of a swap
action~\cite{de-lhoneux-etal-2017-arc}. The Multi-Team tagger performs
morphological tagging~\cite{kirov-etal-2018-unimorph} and lemmatization
jointly; to this end, they use a shared BiLSTM encoder and feed the output of
the tagging as input for the lemmatization, which is predicted as a sequence of
characters.  For efficiency reasons and simplicity, we disabled the use of
external embeddings\footnote{\newcite{ustun-etal-2019-multi} showed that
performance gains from external embeddings are highly complementary to
performance gains from dataset embeddings.} as well as POS embeddings for the
UUParser.

It should be noted that besides the differences in the models and tasks, the
setups also differ among several aspects; the version of UD data (2.3,
2.2)~\cite{nivre-etal-2020-universal}, type of dataset splitting used
(Multi-Team always has train-dev-test), and most interestingly, the dataset
grouping strategies.  ~\newcite{smith-etal-2018-82} manually designed dataset
groups based on typological information, language-relatedness and empirical
evidence;~\newcite{ustun-etal-2019-multi} instead propose pairs:\ every dataset
is matched with one other dataset based on word overlap. For both of the
models, we copy the exact language grouping as in the original
papers\footnote{The full groups can be seen in Appendix~\ref{app:dep} and
~\ref{app:morph}.}.  For comparison of different grouping strategies, we refer
to~\newcite{lin-etal-2019-choosing}.

\begin{table*}
\setlength{\tabcolsep}{3.5pt}
\hspace{-.2cm}
\resizebox{1.005\textwidth}{!}{
\begin{tabular}{l ||  r  | c c c c | c c c c | | r | c c c c}
    \toprule
  &  & \multicolumn{4}{c}{Morphological Tagging (F1)} & \multicolumn{4}{c||}{Lemmatization (Accuracy)} & & \multicolumn{4}{c}{Dependency Parsing (LAS)} \\
Filtering & \#src &  base & concat & gold & pred & base & concat & gold & pred & \#src & base & concat & gold & pred \\
    \midrule
    All & 104 & 92.04 & \cellcolor{red!20} 91.43 & \cellcolor{green!23} 92.75 & \cellcolor{red!6} 91.85 & 91.10 & \cellcolor{red!2} 91.02 & \cellcolor{green!48} 92.55 & \cellcolor{green!10} 91.41 & 58 & 72.92 & \cellcolor{green!38} 74.07 & \cellcolor{green!86} 75.53 & \cellcolor{green!53} 74.52 \\
    \midrule
    Single-lang & 59 & 94.14 & \cellcolor{red!6} 93.94 & \cellcolor{green!56} 95.84 & \cellcolor{red!0} 94.13 & 93.66 & \cellcolor{green!5} 93.83 & \cellcolor{green!69} 95.73 & \cellcolor{green!6} 93.84 & 10 & 80.48 & \cellcolor{red!21} 79.84 & \cellcolor{green!75} 82.74 & \cellcolor{red!6} 80.29 \\
    Multi-lang & 45 & 89.30 & \cellcolor{red!38} 88.14 & \cellcolor{red!20} 88.69 & \cellcolor{red!14} 88.88 & 87.75 & \cellcolor{red!14} 87.33 & \cellcolor{green!20} 88.38 & \cellcolor{green!15} 88.22 & 48 & 71.35 & \cellcolor{green!50} 72.87 & \cellcolor{green!89} 74.03 & \cellcolor{green!65} 73.32 \\
    \bottomrule
\end{tabular}}
\caption{Results per task: overall average, and monolingual vs cross-lingual aggregates. 
\textsc{\#src}:\ number of datasets sources;
\textsc{base}:\ single dataset baseline, \textsc{concat}:\ concatenation of datasets in group, \textsc{gold}:\ gold dataset ids, \textsc{pred}:\ predicted dataset ids. The intensity of colors indicate the difference to the baseline performance.} 
\label{tab:mainResults}
\end{table*}

\subsection{Data Source Prediction}
\label{sec:sourcePrediction}
In this work, we predict data source on the sentence level, because it matches
the language switches at test-time and it improves the accuracy of the
classification.\footnote{However,~\newcite{bhat-etal-2017-joining}
and~\newcite{Vinit:Thesis:2003} have shown the usefulness of word-level
language labels for processing code-switched data.} We use a linear support
vector classifier based on word and character n-grams (without tokenization).
We use this approach here because of simplicity, efficiency and they have shown
to reach competitive performance for text classification
tasks~\cite{zampieri-etal-2017-findings,medvedeva-etal-2017-sparse,coltekin-rama-2018-tubingen,angelo}.
We performed a grid search with $n \in [1-7]$ and all sequential combinations
(1-2, 1-3, etc.) for $n$-grams. For this hyper-parameter tuning, we used the
eight datasets from~\newcite{ustun-etal-2019-multi}, and found the most robust
parameters to be 1-2 for words and 1-5 for characters. The obtained macro
average F1 on all data pairs from~\newcite{ustun-etal-2019-multi} is 95.42, and
on all data groups from~\newcite{smith-etal-2018-82} is 91.76. The performance
difference can be explained by the number of datasets per group, which in the
former setup is always two.  To match the setup during testing, we obtain
predicted dataset identifiers for the training data with 5-fold jack-knifing,
and use these during training.

\section{Results}
\label{sec:results}
We report results for all tasks in two main settings: \textit{in-dataset}, for
setups where we assume that input data is from a distribution present during
training  (\ref{sec:inData}); and \textit{zero-shot}, (\ref{sec:noTrain}), a
setup where this is not the case.  For all reported experiments, we use Labeled
Attachment Score (LAS) for parsing~\cite{zeman-etal-2018-conll}, F1 score for
morphological tagging, and accuracy for lemmatization. We do not perform any
tuning, and thus only report results on development data (if no dev-split is
available we use test). As a control, we compare dataset embeddings to a simple
\textsc{Concat}, training on concatenation of all the data sources from a
dataset group without dataset embeddings.  Reported results are average over 3
runs for the UUparser, for the Multi-Team tagger we did only a single run
because of the computational costs (see Appendix for more details).

\subsection{In-dataset Evaluation}
\label{sec:inData}
The average results over all datasets are shown in Table~\ref{tab:mainResults},
as well as the results for mono-lingual and multi-lingual dataset groups (the
full results can be found in the appendix). These are the takeaways:

\begin{figure}
\centering
\includegraphics[width=\linewidth]{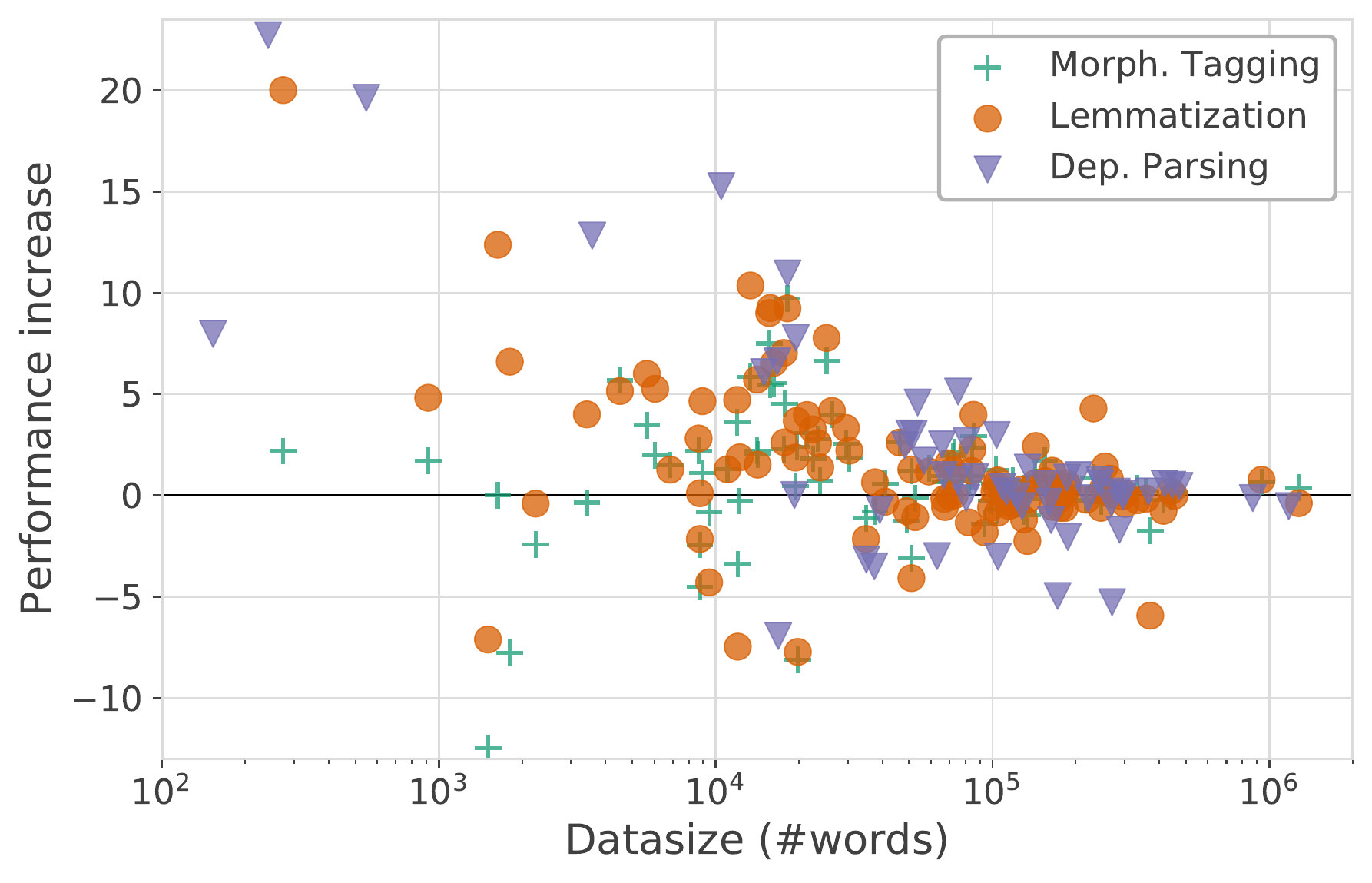}
\caption{Absolute improvement in performance between \textsc{base} and
\textsc{gold} in relation to data size (in number of words in the \textsc{base}
training data) in log scale. Performance difference is absolute, and measured
in the default metric for each task.}
\label{fig:scatter}
\end{figure}

\paragraph{Gold}
Overall, gold dataset embeddings provide substantial gains
(Table~\ref{tab:mainResults}: all). They outperform both \textsc{base} and
\textsc{concat} on all 3 tasks, which confirms previous
findings~\cite{smith-etal-2018-82,ustun-etal-2019-multi}. Gains are largest for
dependency parsing, followed by lemmatization and finally morphological
tagging, where the increase is only 0.71.

\paragraph{Dataset group composition}
Comparing the mono-lingual dataset groups with the multi-lingual groups, we can
see that gold dataset embeddings improve results in both settings for 2/3
tasks.  The only setup where gold dataset embeddings are not beneficial is for
morphological tagging in multi-lingual groups, where the gains for
lemmatization are also only marginal (+0.63 abs.\ compared to base). This may
be attributed to the nature of the tasks, morphological tagging and
lemmatization are more language-specific, making it difficult to transfer
relevant information from another language.  In Figure~\ref{fig:scatter}, we
plot the performance increase from \textsc{base} to using gold dataset ID's in
relation to its dataset size. Unsurprisingly, the largest gains are obtained in
smaller datasets ($<$50,000 words) for all tasks. However, especially for the
morphological tasks, the largest drops are also observed in this range, and
mainly happen for low-resource languages which are paired with a distant
language (e.g. Akkadian (akk\_pisandub) and Irish (ga\_idt)).

\paragraph{Gold vs Predicted}
The \textsc{pred} columns in Table~\ref{tab:mainResults} shows that dataset
embeddings are only beneficial (i.e. outperforming \textsc{base}) for
lemmatization and dependency parsing when we do not have access to the gold
dataset ids but use predicted ids instead.  For lemmatization, the average
increase compared to \textsc{base} is only 0.31 Acc., whereas for dependency
parsing, it is +1.60 LAS.  \textsc{Pred} is mostly beneficial in multi-lingual
dataset groups, which is probably because the performance of the dataset
classifier (Section~\ref{sec:sourcePrediction}) is higher in these cases.

\begin{table}\centering
\resizebox{.7\columnwidth}{!}{
\begin{tabular}{l | r | r r}
    \toprule
    & \#src & concat & pred\\
    \midrule
    All & 53 & 53.80 & \textbf{53.87} \\
    \midrule
    $\exists$ same-lang & 35 & 66.35 & \textbf{66.62} \\
    $\nexists$  same-lang & 18 & \textbf{29.39} & 29.06 \\
    \bottomrule
\end{tabular}}
\caption{LAS scores for zero-shot experiments.}
\label{tab:loo}
\end{table}

\subsection{Zero-shot evaluation}
\label{sec:noTrain}
In many real-world situations, the basic assumption that data instances
originate from the training data distribution is violated, and it becomes
essential to find a good way of using data from other sources, like finding the
best proxy source. To test whether dataset embeddings are still useful in a
zero-shot setup, we perform experiments where we hold out the target dataset
during training and then classify all development sentences into the other
sets. We run this experiment only for dataset groups containing more than 2
datasets (11 groups and 53 datasets, for dependency parsing). As baseline we
compare to a model trained on the concatenation of all other datasets from the
group.  This zero-shot setup is challenging. Results are expected to be overall
lower.

Detailed results are reported in Appendix~\ref{app:dep}--Table~\ref{tab:loo}
summarizes the main results.\footnote{Note that \textsc{base} and \textsc{gold}
are not reported here, because no data from the target dataset is included in
this experiments.} We aggregate over datasets for which another in-language
dataset is available within its dataset group (``$\exists$ same-lang''), and
those where this is not the case. Overall, the performance increase has almost
vanished, having only an 0.07 absolute increase in LAS (`all`). This increase
is void in cases no in-language data exists in the group. Only for datasets for
which a same language dataset exists ($\exists$ same-lang, that is, a treebank
exists for the language but it comes from another distribution/domain), slight
improvements are obtained.  We conclude that dataset embeddings are not useful
in setups when the test instances are from another distribution.

\begin{table}
    \hspace{-.25cm}
    \setlength{\tabcolsep}{3.5pt}
   \resizebox{1.03\columnwidth}{!}{
    \begin{tabular}{l  r r |  r r r r | r r}
    \toprule
                  &          &   &  \multicolumn{4}{c|}{In-dataset training}     & \multicolumn{2}{c}{zero-shot} \\
          dataset  & size & svm & base & concat & gold & pred & concat & pred \\
    \midrule
       nl\_alp & 186k & 0.96 & 84.10 & 84.41 & \textbf{84.97} & 84.74 & 72.03 & \textbf{73.69} \\
      af\_afri & 34k & 1.00 & 79.57 & 78.95 & 79.98 & \textbf{80.44} & \textbf{35.86} & 34.84 \\
      nl\_lassy & 75k & 0.88 & 76.76 & 81.59 & \textbf{81.89} & 81.52 & 74.70 & \textbf{75.48} \\
      de\_gsd & 268k & 1.00 & \textbf{79.76} & 78.96 & 79.39 & 79.35 & 14.30 & \textbf{15.09} \\
    \midrule
       en\_pud & 0 & 0.00 & - & - & - & - & \textbf{81.90} & 81.89 \\
      en\_ewt & 205k & 0.91 & 82.43 & 82.60 & \textbf{83.42} & 82.82 & 71.11 & \textbf{71.23} \\
     en\_lines & 50k & 0.77 & 76.15 & 75.06 & \textbf{79.20} & 77.14 & \textbf{74.71} & 74.68 \\
      en\_gum & 54k & 0.71 & 78.18 & 80.32 & \textbf{82.77} & 80.58 & \textbf{80.12} & 79.98 \\
    \midrule
    \end{tabular}
    }
    \caption{Full results for the \textsc{af-de-nl} and \textsc{en} dataset group (LAS). \textsc{size} refers to training size in number of words. \textsc{svm} accuracy of SVM language predictions. }
    \label{tab:examples}
\end{table}

For demonstration purposes, we highlight the full results of two dataset groups
in Table~\ref{tab:examples}. The first (multi-lingual) group shows that
dataset-embeddings are mainly beneficial for languages with multiple datasets,
both in the in-domain and zero-shot setting. In this particular dataset group,
prediction of the embeddings performs on-par with the gold labels, probably
because of the high performance of the classifier. In contrast, in the second
(mono-lingual) group, the classifier scores lower, and \textsc{pred} prediction
performances are lower compared to \textsc{gold}. For this group, predicted
dataset embeddings are outperformed by a simple dataset concatenation.

\section{Conclusion}
We provide an extensive evaluation of dataset embeddings in two large-scale
settings where they were used
successfully~\cite{smith-etal-2018-82,ustun-etal-2019-multi}. In setups where
in-distribution training data is available, we found dataset embeddings more
useful in monolingual dataset groups, compared to cross-lingual ones. In
general, performance gains were the largest for 1) datasets for which another
same-language datasets was available during training 2) small datasets 3)
datasets which were part of a large dataset group.  However, with predicted
id's, their benefit is limited, contrary to gold information.  When moving to
zero-shot setups, the performance increases become negligible (except for some
particular datasets).  In particular, without in-source training data, dataset
embeddings work in some cases when another treebank for the language exists;
but this gain is not consistent and often small.  Overall, we find dataset
embeddings fail to be a viable adaptation method when no in-source data is
available. Hence, in many realistic out-of-distribution setups, their benefit
vanishes.

\section*{Acknowledgements}
We would like to thank Gertjan van Noord for feedback on an earlier version of
this paper. We thank NVIDIA, the HPC cluster at the ITU and the University of
Groningen for computing resources.  This research was supported by an Amazon
Research Award, an STSM in the Multi3Generation COST action (CA18231), and
grant 9063-00077B (Danmarks Frie Forskningsfond).

\bibliography{papers}
\bibliographystyle{acl_natbib}

\newpage
\appendix
\section{Reproducability report}
The UUParser was run on two E5-2660 v3's (40 threads total, we used only 30),
and took on average approximately 20 hours on a single thread per model.  For
three random seeds, 16 dataset groups, and approximately 5 settings (4 from
Section~\ref{sec:inData} +1 from  Section~\ref{sec:noTrain} were we only used
half of the groups in two settings), the total number of models is 240. So the
total computation walltime was 160 hours (approximately a week).

The Multi-Team tagger was run on two Tesla V100's (one model per V100, so two
models were trained in parallel). On average this took approximately 3 hours.
For this setup we had 104 dataset pairs, of which 82 were unique (if the pairs
consist of the same two languages, we only trained one model). For the
Multi-Team tagger, we ran four setups (Section~\ref{sec:inData}), totaling to
328 models. The total computation time was 984 hours, which divided by 2 gpus
resulted in a walltime of 492 hours (approximately three weeks).

Regarding the differences in the settings (\textsc{base}, \textsc{concat},
\textsc{gold} and \textsc{pred}), there were no clear trends in differences in
run-time, even the \textsc{base} settings (where multiple models where trained
for 1 dataset group) was equal in runtime compared to the other settings where
one large model was trained.

It should be noted that the UUParser uses a maximum of 15,000 sentences per
epoch, and the Multi-Team tagger 500,000 words (default settings), which makes
training times substantially shorter (especially for the UUParser), and reduced
the memory usage. Excluding external embeddings helped us reduce the running
time and memory usage even further. For the UUParser, a maximum of 8GB ram is
enough for training a single model (~4GB on average), and the Multi-Team tagger
requires a minimum of 8GB of GPU RAM.

For all the other settings and hyperparameters, we exactly replicated the
original code from the ~\newcite{smith-etal-2018-82} and
~\newcite{ustun-etal-2019-multi}, and thus refer to their papers for
experimental details. The only adaptation we made to the systems is that for
the UUParser we added support for supplying the dataset information in the
connlu misc-column (the adapted version is available in our repo).

\section{Aggregates over all results}
For easier analysis, we provide average scores over aggregates of datasets. To
this end, we propose to use data-filters, and report average scores over
specific subsets of the data. The results are shown in Table~\ref{tab:filters}.
The filters show aggregates over a) whether the training portion of the dataset
is small ($<$ 30,000 words) or large b) whether the dataset group to which this
dataset belongs is mono-lingual or multi-lingual c) whether another dataset
with the same language is available in the dataset group d) whether the svm
classifier predicts the datasource id's with an accuracy of $>$95\% accuracy e)
whether the word overlap is larger then 10\%.

\begin{table*}
\resizebox{\textwidth}{!}{
\begin{tabular}{l r r r r | r r r r | r r r r | | r r r | r r r r}
    \toprule
  & \multicolumn{4}{c}{Language Pairs} & \multicolumn{4}{c}{Morphological Tagging (F1)} & \multicolumn{4}{c||}{Lemmatization (Accuracy)} & \multicolumn{3}{c}{Language Clusters} & \multicolumn{4}{c}{Dependency Parsing (LAS)} \\
Filtering & \#sets & size & WO & svm & base & concat & gold & pred & base & concat & gold & pred & \#sets & size & svm & base & concat & gold & pred \\
    \midrule
    All & 104 & 121 & 0.31 & 0.95 & 92.04 & 91.43 & 92.75 & 91.85 & 91.10 & 91.02 & 92.55 & 91.41 & 58 & 161 & 0.91 & 72.92 & 74.07 & 75.53 & 74.52 \\
    \midrule
    Large & 65 & 186 & 0.28 & 0.94 & 95.40 & 94.78 & 95.72 & 94.98 & 95.92 & 95.08 & 96.00 & 95.19 & 46 & 200 & 0.92 & 80.80 & 79.62 & 80.98 & 80.06 \\
    Small & 39 & 13 & 0.35 & 0.98 & 86.46 & 85.85 & 87.78 & 86.65 & 83.07 & 84.24 & 86.79 & 85.11 & 12 & 10 & 0.88 & 42.75 & 52.79 & 54.65 & 53.30 \\
    \midrule
    Multi-lang & 45 & 77 & 0.13 & 0.99 & 89.30 & 88.14 & 88.69 & 88.88 & 87.75 & 87.33 & 88.38 & 88.22 & 48 & 163 & 0.92 & 71.35 & 72.87 & 74.03 & 73.32 \\
    Single-lang & 59 & 154 & 0.44 & 0.92 & 94.14 & 93.94 & 95.84 & 94.13 & 93.66 & 93.83 & 95.73 & 93.84 & 10 & 149 & 0.86 & 80.48 & 79.84 & 82.74 & 80.29 \\
    \midrule
    $\exists$ same-lang & 59 & 154 & 0.44 & 0.92 & 94.14 & 93.94 & 95.84 & 94.13 & 93.66 & 93.83 & 95.73 & 93.84 & 35 & 187 & 0.86 & 77.32 & 77.33 & 79.29 & 77.76 \\
    $\nexists$ same-lang & 45 & 77 & 0.13 & 0.99 & 89.30 & 88.14 & 88.69 & 88.88 & 87.75 & 87.33 & 88.38 & 88.22 & 23 & 121 & 0.98 & 66.23 & 69.11 & 69.80 & 69.60 \\
    \midrule
    pred$<$95\% & 33 & 177 & 0.41 & 0.87 & 95.15 & 94.05 & 96.45 & 94.24 & 94.57 & 94.12 & 95.77 & 94.24 & 24 & 134 & 0.79 & 73.60 & 75.43 & 78.23 & 76.00 \\
    pred$>$95\% & 71 & 95 & 0.26 & 0.99 & 90.60 & 90.21 & 91.02 & 90.75 & 89.49 & 89.57 & 91.05 & 90.10 & 34 & 179 & 0.99 & 72.45 & 73.11 & 73.62 & 73.48 \\
    \midrule
    highWO & 78 & 139 & 0.39 & 0.94 & 93.22 & 92.98 & 94.69 & 93.36 & 92.82 & 93.11 & 94.94 & 93.38 & 58 & 161 & 0.91 & 72.92 & 74.07 & 75.53 & 74.52 \\
    lowWO & 26 & 67 & 0.05 & 1.00 & 88.51 & 86.79 & 86.91 & 87.34 & 85.94 & 84.72 & 85.38 & 85.49 \\
    \bottomrule
    \end{tabular}
}
\caption{Results per tasks, with averages over different dataset filters. \textsc{\#sets}:\ number of datasets, \textsc{size}:\ training data size of dataset (in 1,000 words), 
\textsc{svm}:\ F1 score of dataset classifier, \textsc{base}:\ single dataset baseline, \textsc{concat}:\ concatenation of datasets in cluster, \textsc{gold}:\ gold dataset ids, \textsc{pred}:\ predicted dataset ids.}
\label{tab:filters}
\end{table*}

\section{Full results for dependency parsing}
\label{app:dep}
Table~\ref{tab:fullDep} shows the results of the
UUParser~\cite{smith-etal-2018-82} for each dataset, grouped by dataset groups.
All results are the average over three runs. We do not report scores for
datasets without in-source training data in the `parser setting' columns (which
corresponds to Section~\ref{sec:inData} of the paper), as training data is
necessary for those settings. 

For the `without train' columns (corresponding to Section~\ref{sec:noTrain} of
the paper), we do not include results for dataset groups of size two; this is
because we leave one training set out, and try to predict for the corresponding
development set in which set it belongs. For groups of size two, this
classification is trivial and non-informative, as there is only one dataset
left. The left-out datasets are not taken into account for the averages. The
reported results are on development splits, except for datasets which did not
have a development split available, there we used test (indicated with * in the
table) as we do not perform any tuning.

\begin{table*}
\centering
\renewcommand{\arraystretch}{.9}
\resizebox{.674\textwidth}{!}{
\begin{tabular}{l l  r r |  r r r r | r r}
            &         &          &     & \multicolumn{4}{c|}{In-dataset training} & \multicolumn{2}{c}{Zero-shot} \\
    cluster & dataset & size & svm & base & concat & gold & pred & concat & pred \\
    \midrule
af-de-nl      & nl\_alpino & 186,046 & 0.96 & 84.10 & 84.41 & \textbf{84.97} & 84.74 & 72.03 & \textbf{73.69} \\
     & af\_afribooms & 33,894 & 1.00 & 79.57 & 78.95 & 79.98 & \textbf{80.44} & \textbf{35.86} & 34.84 \\
     & nl\_lassysmall & 75,134 & 0.88 & 76.76 & 81.59 & \textbf{81.89} & 81.52 & 74.70 & \textbf{75.48} \\
     & de\_gsd & 268,414 & 1.00 & \textbf{79.76} & 78.96 & 79.39 & 79.35 & 14.30 & \textbf{15.09} \\
    \midrule
e-sla      & uk\_iu & 75,098 & 0.99 & 79.86 & 79.49 & \textbf{80.63} & 80.53 & 35.69 & \textbf{35.99} \\
     & ru\_taiga & 10,479 & 0.35 & 56.47 & 69.32 & \textbf{71.73} & 69.30 & \textbf{68.81} & \textbf{68.81} \\
     & ru\_syntagrus & 871,521 & 0.99 & \textbf{87.26} & 87.07 & 87.13 & 87.24 & \textbf{59.52} & 59.34 \\
    \midrule
en      & en\_pud & 0 & 0.00 & - & - & - & - & \textbf{81.90} & 81.89 \\
     & en\_ewt & 204,607 & 0.91 & 82.43 & 82.60 & \textbf{83.42} & 82.82 & 71.11 & \textbf{71.23} \\
     & en\_lines & 50,096 & 0.77 & 76.15 & 75.06 & \textbf{79.20} & 77.14 & \textbf{74.71} & 74.68 \\
     & en\_gum & 53,686 & 0.71 & 78.18 & 80.32 & \textbf{82.77} & 80.58 & \textbf{80.12} & 79.98 \\
    \midrule
es-ca      & ca\_ancora & 418,494 & 1.00 & 87.97 & 88.41 & \textbf{88.55} & 88.51 & - & - \\
     & es\_ancora & 446,145 & 1.00 & 87.44 & 87.74 & 87.97 & \textbf{88.07} & - & - \\
    \midrule
finno      & et\_edt & 287,859 & 1.00 & \textbf{79.48} & 77.47 & 77.79 & 77.66 & \textbf{14.38} & 14.32 \\
     & fi\_tdt & 162,827 & 0.77 & \textbf{79.48} & 71.43 & 78.27 & 70.45 & 47.21 & \textbf{47.44} \\
     & fi\_ftb & 127,845 & 0.69 & \textbf{79.58} & 70.48 & 79.05 & 70.88 & 51.46 & \textbf{52.16} \\
     & sme\_giella & 16,835 & 1.00 & \textbf{63.17} & 53.08 & 56.23 & 55.32 & \textbf{6.58} & 6.45 \\
     & fi\_pud & 0 & 0.00 & - & - & - & - & 74.58 & \textbf{74.94} \\
    \midrule
fr      & fr\_spoken & 14,952 & 0.93 & 71.39 & 76.15 & \textbf{77.48} & 76.72 & \textbf{53.50} & 53.35 \\
     & fr\_gsd & 366,372 & 0.96 & 88.23 & 88.09 & \textbf{88.43} & 88.30 & 75.65 & \textbf{76.36} \\
     & fr\_sequoia & 51,906 & 0.69 & 85.72 & 85.35 & \textbf{88.75} & 87.13 & 80.62 & \textbf{80.78} \\
    \midrule
indic      & ur\_udtb & 108,690 & 1.00 & 78.15 & 78.43 & \textbf{78.58} & \textbf{78.58} & - & - \\
     & hi\_hdtb & 281,057 & 1.00 & 89.20 & 89.27 & \textbf{89.38} & \textbf{89.38} & - & - \\
    \midrule
iranian      & fa\_seraji & 122,180 & 1.00 & 82.45 & 82.26 & 82.41 & \textbf{82.48} & - & - \\
     & kmr\_mg & 242 & 0.99 & 12.24 & 34.76 & 34.96 & \textbf{35.38} & - & - \\
    \midrule
it      & it\_isdt & 294,397 & 0.99 & 87.71 & 87.58 & 87.78 & \textbf{87.89} & - & - \\
     & it\_postwita & 103,553 & 0.98 & 75.17 & 77.72 & \textbf{78.15} & 77.83 & - & - \\
    \midrule
ko      & ko\_gsd & 56,687 & 0.68 & 76.70 & 65.62 & \textbf{78.42} & 63.56 & - & - \\
     & ko\_kaist & 296,446 & 0.95 & \textbf{83.08} & 79.90 & 83.03 & 80.93 & - & - \\
    \midrule
n-ger      & no\_nynorsklia & 3,583 & 0.90 & 50.05 & 62.27 & 62.87 & \textbf{62.91} & 52.89 & \textbf{53.27} \\
     & fo\_oft & 0 & 0.00 & - & - & - & - & 39.57 & \textbf{40.87} \\
     & sv\_talbanken & 66,673 & 0.94 & 77.37 & 76.35 & \textbf{78.37} & 77.57 & 70.49 & \textbf{71.97} \\
     & no\_bokmaal & 243,887 & 0.97 & 87.21 & 87.67 & \textbf{87.97} & 87.76 & \textbf{76.79} & 76.12 \\
     & sv\_pud & 0 & 0.00 & - & - & - & - & \textbf{77.89} & 77.65 \\
     & sv\_lines & 48,325 & 0.91 & 76.48 & 77.71 & \textbf{78.95} & 78.37 & 71.66 & \textbf{72.18} \\
     & no\_nynorsk & 245,330 & 0.98 & 85.67 & 85.49 & \textbf{86.27} & 85.93 & 73.50 & \textbf{74.36} \\
     & da\_ddt & 80,378 & 0.97 & \textbf{76.97} & 73.79 & 76.84 & 76.06 & 52.04 & \textbf{52.27} \\
    \midrule
old      & cu\_proiel & 37,432 & 1.00 & \textbf{76.62} & 73.91 & 73.12 & 74.22 & \textbf{5.36} & 4.95 \\
     & got\_proiel & 35,024 & 1.00 & \textbf{71.46} & 69.02 & 68.29 & 69.06 & \textbf{8.53} & 8.25 \\
     & grc\_proiel & 187,049 & 1.00 & \textbf{76.09} & 74.32 & 74.03 & 74.35 & \textbf{53.90} & 53.51 \\
     & la\_perseus & 18,184 & 0.88 & 42.55 & 50.45 & \textbf{53.51} & 52.61 & \textbf{42.61} & 41.12 \\
     & la\_proiel & 171,928 & 0.99 & \textbf{71.18} & 67.03 & 66.22 & 66.94 & 42.68 & \textbf{43.68} \\
     & grc\_perseus & 159,895 & 1.00 & \textbf{61.78} & 61.46 & 61.46 & 61.36 & 46.42 & \textbf{47.87} \\
     & la\_ittb & 270,403 & 1.00 & \textbf{79.40} & 74.29 & 74.14 & 74.69 & \textbf{41.66} & 40.61 \\
    \midrule
pt-gl      & gl\_ctg & 86,676 & 0.95 & 80.21 & 79.85 & \textbf{81.11} & 80.56 & 63.78 & \textbf{64.19} \\
     & pt\_bosque & 222,069 & 1.00 & \textbf{87.68} & 87.11 & 87.54 & 87.59 & 49.85 & \textbf{50.26} \\
     & gl\_treegal & 16,707 & 0.62 & 69.15 & 67.24 & \textbf{75.76} & 69.03 & 61.62 & \textbf{61.71} \\
    \midrule
sw-sla      & sl\_sst & 19,473 & 0.96 & 58.65 & 65.65 & \textbf{66.42} & 66.31 & \textbf{52.06} & 51.81 \\
     & sr\_set & 65,764 & 0.86 & 83.91 & 84.07 & \textbf{86.42} & 85.91 & 75.63 & \textbf{75.66} \\
     & hr\_set & 154,055 & 0.94 & 80.66 & 80.22 & \textbf{81.23} & 81.07 & 65.74 & \textbf{66.08} \\
     & sl\_ssj & 112,530 & 0.99 & 85.27 & 84.89 & \textbf{85.46} & 85.23 & 65.28 & \textbf{65.83} \\
    \midrule
turkic      & ug\_udt & 19,262 & 1.00 & 61.43 & 60.86 & \textbf{61.45} & 60.88 & 1.88 & \textbf{2.80} \\
     & bxr\_bdt & 153 & 0.98 & 9.95 & \textbf{17.99} & 17.92 & 17.04 & 4.85 & \textbf{5.37} \\
     & tr\_imst & 39,169 & 1.00 & \textbf{57.01} & 55.51 & 56.29 & 56.63 & \textbf{9.51} & 9.38 \\
     & kk\_ktb & 547 & 1.00 & 11.54 & 30.52 & \textbf{31.16} & 29.14 & \textbf{7.03} & 6.19 \\
    \midrule
w-sla      & sk\_snk & 80,575 & 0.98 & 80.39 & 82.49 & \textbf{83.07} & 82.51 & \textbf{59.77} & 59.57 \\
     & cs\_pud & 0 & 0.00 & - & - & - & - & \textbf{83.79} & 83.77 \\
     & cs\_pdt & 1,175,374 & 0.92 & \textbf{87.92} & 87.41 & 87.40 & 87.39 & 79.07 & \textbf{79.33} \\
     & pl\_sz & 63,070 & 0.46 & \textbf{85.31} & 80.88 & 82.31 & 81.49 & 67.40 & \textbf{67.88} \\
     & hsb\_ufal & 460 & 0.90 & 6.40 & 45.24 & \textbf{46.30} & 44.99 & \textbf{39.10} & 38.55 \\
     & pl\_lfg & 104,750 & 0.71 & \textbf{90.98} & 86.51 & 87.96 & 87.53 & \textbf{70.76} & 70.49 \\
     & cs\_fictree & 134,059 & 0.78 & 85.77 & 86.92 & \textbf{87.14} & 87.08 & \textbf{83.75} & 82.91 \\
     & cs\_cac & 473,622 & 0.83 & 86.86 & \textbf{87.40} & 87.32 & \textbf{87.40} & \textbf{83.73} & 83.57 \\
    \bottomrule
\end{tabular}}
\caption{LAS scores on all development splits of the dependency parser. *: datasets for which no development data was available, we report results on test data instead. The `in-dataset setting' results corresponds to Section~\ref{sec:results}, and `zero-shot' to Section~\ref{sec:noTrain}, where we assume no in-source training data. \textsc{size}: size of training data in words. \textsc{svm}: F1 score of svm classifier on dataset prediction. \textsc{base}: single dataset baseline. \textsc{concat}: concatenation of datasets. \textsc{gold}: gold dataset embeddings. \textsc{pred}: predicted dataset embeddings.}
\label{tab:fullDep}
\end{table*}

\section{Full results for morphological tagging and lemmatization}
\label{app:morph}
Table~\ref{tab:fullMorph} shows the results of the Multi-Team
tagger~\cite{ustun-etal-2019-multi} on the development data for each dataset.
Because of the computational costs, results are over a single run.  The second
column shows the `help-dataset' that each dataset is paired with, based on word
overlap.

Note that data sizes are different compared to Table~\ref{tab:fullDep} due to a
re-split of the data by~\cite{mccarthy-etal-2019-sigmorphon}, and different UD
versions (~\newcite{ustun-etal-2019-multi} used 2.3 whereas
~\newcite{smith-etal-2018-82} used 2.2). Another effect of this re-split is
that for all datasets, a train, development and test split is available. Also
note that dataset prediction (\textsc{svm}) scores reported are on the train
data; so if the score is 1.00, \textsc{pred} and \textsc{gold} can still have
different scores because the dataset prediction was not equally accurate on the
development data.

\begin{table*}
\centering
\resizebox{\textwidth}{!}{
\begin{tabular}{l l r r r | r r r r | r r r r}
\toprule
    & & & & & \multicolumn{4}{c}{Morphological Tagging (F1)} & \multicolumn{4}{c}{Lemmatization (Accuracy)} \\
    dataset & additional & size & \textsc{wo} & svm & base & concat & gold & pred & base & concat & gold & pred \\
    \midrule
    af\_afribooms & nl\_alpino & 40,390 & 0.20 & 99.85 & 96.50 & 96.42 & \textbf{96.96} & 96.87 & 95.25 & 94.97 & 96.01 & \textbf{96.93} \\
    akk\_pisandub & cs\_pdt & 1,505 & 0.02 & 100.00 & \textbf{81.96} & 67.99 & 69.49 & 71.96 & \textbf{41.33} & 34.67 & 34.22 & 35.11 \\
    ar\_padt & ar\_pud & 231,625 & 0.18 & 96.64 & 95.29 & 95.46 & \textbf{96.17} & 95.44 & 90.79 & 91.52 & \textbf{95.08} & 91.80 \\
    ar\_pud & ar\_padt & 17,645 & 0.56 & 96.64 & 89.97 & 87.49 & \textbf{92.24} & 87.20 & 77.36 & 62.83 & \textbf{84.39} & 57.39 \\
    be\_hse & ru\_syntagrus & 6,855 & 0.08 & 99.98 & 80.51 & 80.61 & 82.01 & \textbf{83.49} & 78.48 & 75.67 & 79.75 & \textbf{81.58} \\
    bg\_btb & ru\_syntagrus & 133,659 & 0.12 & 99.60 & \textbf{97.85} & 96.67 & 96.89 & 96.97 & \textbf{96.95} & 94.54 & 94.70 & 94.95 \\
    bm\_crb & cs\_pdt & 12,025 & 0.09 & 99.98 & \textbf{94.03} & 89.49 & 90.64 & 89.92 & \textbf{87.86} & 78.40 & 80.40 & 80.09 \\
    br\_keb & no\_bokmaal & 8,772 & 0.07 & 99.87 & \textbf{90.89} & 90.53 & 88.45 & 89.67 & 88.75 & \textbf{89.77} & 88.86 & 88.35 \\
    bxr\_bdt & ru\_syntagrus & 8,770 & 0.04 & 99.94 & \textbf{83.46} & 80.65 & 78.94 & 79.96 & \textbf{82.71} & 80.65 & 80.56 & 81.63 \\
    ca\_ancora & es\_ancora & 441,014 & 0.18 & 99.71 & \textbf{98.61} & 98.48 & 98.57 & 98.56 & 98.35 & 98.42 & 98.56 & \textbf{98.69} \\
    cs\_cac & cs\_pdt & 414,810 & 0.57 & 90.62 & \textbf{97.19} & 96.71 & 96.98 & 96.72 & \textbf{98.03} & 97.05 & 97.27 & 97.24 \\
    cs\_cltt & cs\_pdt & 29,549 & 0.81 & 99.87 & 94.62 & 96.65 & \textbf{97.16} & 96.62 & 94.02 & \textbf{97.56} & 97.35 & \textbf{97.56} \\
    cs\_fictree & cs\_pdt & 143,508 & 0.58 & 95.01 & 95.89 & 94.89 & \textbf{96.54} & 95.36 & 95.21 & 97.07 & \textbf{97.65} & 97.15 \\
    cs\_pdt & cs\_cac & 1,278,252 & 0.27 & 90.62 & 96.65 & 96.78 & \textbf{97.05} & 96.76 & \textbf{97.50} & 97.10 & 97.12 & 97.19 \\
    cs\_pud & cs\_pdt & 15,614 & 0.79 & 98.98 & 87.38 & 96.26 & 94.88 & \textbf{96.38} & 87.03 & 96.84 & 96.03 & \textbf{97.00} \\
    cu\_proiel & ru\_syntagrus & 50,963 & 0.04 & 99.98 & \textbf{94.71} & 92.94 & 91.62 & 92.12 & \textbf{95.17} & 92.66 & 91.09 & 91.28 \\
    da\_ddt & no\_bokmaal & 85,373 & 0.25 & 97.38 & 92.58 & 95.21 & \textbf{95.52} & 95.26 & 92.27 & 95.60 & \textbf{96.25} & 95.30 \\
    de\_gsd & fr\_gsd & 246,633 & 0.06 & 99.94 & \textbf{93.48} & 92.18 & 93.16 & 93.05 & \textbf{96.55} & 94.95 & 95.94 & 95.60 \\
    el\_gdt & grc\_proiel & 52,583 & 0.04 & 99.96 & \textbf{96.51} & 96.25 & 96.36 & 96.45 & 95.00 & 94.52 & 93.93 & \textbf{95.15} \\
    en\_ewt & en\_gum & 218,154 & 0.30 & 89.27 & 95.73 & \textbf{95.75} & 95.49 & 95.32 & \textbf{97.02} & 96.79 & 96.81 & 96.46 \\
    en\_gum & en\_ewt & 67,381 & 0.57 & 89.27 & 94.45 & 94.39 & \textbf{95.11} & 94.29 & \textbf{96.94} & 93.94 & 96.37 & 94.63 \\
    en\_lines & en\_ewt & 70,079 & 0.56 & 91.20 & 95.14 & 93.33 & \textbf{95.88} & 94.71 & \textbf{97.39} & 94.98 & 97.34 & 96.59 \\
    en\_partut & en\_ewt & 40,974 & 0.63 & 95.66 & 93.45 & 90.75 & \textbf{94.03} & 91.83 & \textbf{97.58} & 96.53 & 97.27 & 95.83 \\
    en\_pud & en\_ewt & 17,727 & 0.68 & 95.14 & 90.87 & 95.08 & \textbf{95.40} & 94.91 & 93.55 & 95.62 & \textbf{96.17} & 94.68 \\
    es\_ancora & es\_gsd & 454,069 & 0.47 & 87.10 & 98.34 & 97.46 & \textbf{98.48} & 97.97 & \textbf{98.60} & 97.15 & 98.59 & 97.80 \\
    es\_gsd & es\_ancora & 358,355 & 0.42 & 87.10 & 97.35 & 96.15 & \textbf{97.55} & 96.97 & \textbf{98.59} & 95.81 & 98.43 & 97.65 \\
    et\_edt & cs\_pdt & 371,564 & 0.02 & 99.75 & \textbf{96.60} & 94.65 & 94.86 & 94.68 & \textbf{94.73} & 89.62 & 88.80 & 89.09 \\
    eu\_bdt & es\_ancora & 104,530 & 0.05 & 99.96 & \textbf{95.10} & 93.62 & 94.41 & 94.38 & \textbf{96.32} & 95.35 & 95.47 & 95.52 \\
    fa\_seraji & ur\_udtb & 127,371 & 0.12 & 100.00 & 97.15 & 97.24 & 97.26 & \textbf{97.31} & 95.00 & 94.24 & \textbf{95.29} & 94.97 \\
    fi\_ftb & fi\_tdt & 142,514 & 0.37 & 71.46 & 95.25 & 94.94 & \textbf{96.18} & 94.94 & 92.15 & 90.99 & \textbf{92.75} & 90.85 \\
    fi\_pud & fi\_tdt & 13,356 & 0.49 & 94.58 & 91.56 & 96.97 & \textbf{97.41} & 96.81 & 78.62 & 86.06 & \textbf{88.98} & 86.78 \\
    fi\_tdt & fi\_ftb & 173,899 & 0.30 & 71.46 & 96.54 & 95.32 & \textbf{97.07} & 95.18 & \textbf{92.42} & 90.27 & 92.22 & 90.03 \\
    fo\_oft & no\_nynorsk & 8,960 & 0.12 & 99.83 & 90.36 & 86.49 & \textbf{91.46} & 90.38 & 83.87 & 81.59 & \textbf{88.52} & 86.66 \\
    fr\_gsd & fr\_sequoia & 333,477 & 0.15 & 93.23 & 97.71 & 97.65 & \textbf{98.07} & 97.40 & \textbf{97.74} & 96.52 & 97.50 & 96.14 \\
    fr\_partut & fr\_gsd & 23,443 & 0.82 & 97.28 & 94.72 & 96.33 & \textbf{97.51} & 96.13 & 94.20 & 95.13 & \textbf{96.78} & 95.10 \\
    fr\_sequoia & fr\_gsd & 58,963 & 0.66 & 93.23 & 96.74 & 96.79 & \textbf{98.07} & 96.21 & 96.99 & 96.90 & \textbf{98.17} & 95.81 \\
    fr\_spoken & fr\_gsd & 30,410 & 0.77 & 99.33 & 95.81 & 97.10 & \textbf{97.64} & 97.44 & 96.77 & 98.79 & 98.98 & \textbf{99.07} \\
    ga\_idt & cs\_pdt & 19,812 & 0.04 & 99.95 & \textbf{83.38} & 76.03 & 75.27 & 77.94 & \textbf{84.63} & 76.48 & 76.91 & 79.01 \\
    gl\_ctg & es\_ancora & 114,228 & 0.40 & 99.88 & 97.33 & 97.22 & \textbf{97.38} & 97.36 & 98.12 & 98.14 & 98.16 & \textbf{98.37} \\
    gl\_treegal & gl\_ctg & 21,366 & 0.53 & 92.26 & 91.56 & 82.76 & \textbf{94.64} & 84.77 & 92.69 & 94.69 & \textbf{96.66} & 95.57 \\
    got\_proiel & no\_nynorsk & 48,980 & 0.01 & 99.91 & \textbf{95.20} & 94.61 & 93.97 & 93.94 & \textbf{95.35} & 94.60 & 94.58 & 94.23 \\
    grc\_perseus & grc\_proiel & 173,299 & 0.25 & 99.95 & 94.86 & 94.74 & \textbf{95.05} & \textbf{95.05} & \textbf{93.24} & 92.67 & 93.15 & 93.19 \\
    grc\_proiel & grc\_perseus & 185,142 & 0.31 & 99.95 & 96.92 & 96.98 & 97.08 & \textbf{97.11} & 95.85 & 95.90 & \textbf{96.46} & 96.30 \\
    he\_htb & ru\_gsd & 134,397 & 0.00 & 100.00 & \textbf{96.26} & 96.17 & 96.07 & 96.23 & 96.62 & 96.52 & 96.44 & \textbf{96.71} \\
    hi\_hdtb & mr\_ufal & 295,265 & 0.01 & 100.00 & 96.70 & 96.87 & 96.77 & \textbf{96.89} & \textbf{98.57} & 98.34 & 98.52 & 98.40 \\
    hr\_set & sr\_set & 164,557 & 0.28 & 88.98 & 95.14 & 94.85 & \textbf{95.50} & 94.59 & \textbf{95.81} & 94.53 & 95.25 & 94.52 \\
    hsb\_ufal & cs\_pdt & 9,475 & 0.08 & 99.95 & \textbf{79.92} & 77.49 & 79.09 & 76.84 & \textbf{82.68} & 74.70 & 78.39 & 76.76 \\
    hu\_szeged & et\_edt & 34,903 & 0.03 & 99.95 & \textbf{92.27} & 91.05 & 91.15 & 89.75 & \textbf{90.09} & 88.37 & 87.94 & 84.66 \\
    hy\_armtdp & ru\_pud & 19,419 & 0.00 & 100.00 & 90.34 & \textbf{91.31} & 90.81 & 91.13 & 90.00 & \textbf{92.15} & 91.87 & 92.01 \\
    id\_gsd & es\_gsd & 101,687 & 0.11 & 99.92 & 92.72 & 92.87 & 93.30 & \textbf{93.39} & 98.77 & 98.87 & \textbf{98.97} & 98.81 \\
        it\_isdt & it\_partut & 250,714 & 0.28 & 79.42 & 98.07 & 97.91 & \textbf{98.26} & 97.77 & 97.52 & 96.70 & \textbf{97.87} & 96.85 \\
    it\_partut & it\_isdt & 46,228 & 0.94 & 79.42 & 96.11 & 98.47 & \textbf{98.73} & 98.30 & 96.07 & 97.78 & \textbf{98.68} & 97.21 \\
    it\_postwita & it\_isdt & 104,437 & 0.46 & 98.49 & 95.76 & 96.13 & 96.15 & \textbf{96.53} & 94.15 & \textbf{96.14} & 94.89 & 95.20 \\
    it\_pud & it\_isdt & 19,634 & 0.69 & 94.02 & 94.30 & 85.62 & \textbf{96.69} & 84.00 & 93.32 & 96.64 & \textbf{97.00} & 95.22 \\
    \end{tabular}}
    \end{table*}

    \begin{table*}
    \centering
    \resizebox{\textwidth}{!}{
    \begin{tabular}{l l r r r | r r r r | r r r r}
    ja\_gsd & ja\_pud & 154,453 & 0.14 & 92.59 & 94.59 & 96.10 & \textbf{96.31} & 95.96 & 98.06 & \textbf{98.81} & 98.77 & 98.77 \\
    ja\_modern & ja\_gsd & 12,213 & 0.29 & 99.78 & \textbf{95.93} & 95.69 & 95.65 & 95.86 & 93.67 & 95.96 & 95.55 & \textbf{96.59} \\
    ja\_pud & ja\_gsd & 22,450 & 0.64 & 92.59 & 95.85 & \textbf{97.91} & 97.66 & 97.29 & 96.08 & 99.30 & \textbf{99.34} & 99.01 \\
    kmr\_mg & es\_gsd & 8,680 & 0.03 & 100.00 & 86.43 & 86.93 & \textbf{88.64} & 87.61 & 88.38 & 90.72 & \textbf{91.19} & 90.44 \\
    ko\_gsd & ko\_kaist & 69,382 & 0.33 & 92.29 & 93.53 & 86.94 & \textbf{94.97} & 87.35 & 89.88 & 89.30 & \textbf{91.47} & 85.59 \\
    ko\_kaist & ko\_gsd & 302,384 & 0.12 & 92.29 & 95.86 & 95.48 & \textbf{95.97} & 95.39 & \textbf{94.30} & 94.20 & 93.88 & 92.37 \\
    ko\_pud & ko\_kaist & 14,106 & 0.55 & 97.97 & 93.41 & 82.62 & \textbf{95.62} & 83.29 & 92.36 & 75.41 & \textbf{98.08} & 75.41 \\
    kpv\_ikdp & ru\_syntagrus & 916 & 0.26 & 99.95 & 61.38 & 48.34 & \textbf{63.10} & 55.79 & 56.63 & 55.42 & \textbf{61.45} & \textbf{61.45} \\
    kpv\_lattice & ru\_syntagrus & 1,805 & 0.09 & 99.96 & \textbf{75.26} & 66.32 & 67.50 & 67.36 & 57.69 & 58.24 & \textbf{64.29} & 61.54 \\
    la\_ittb & la\_proiel & 298,460 & 0.37 & 99.98 & 97.08 & 97.23 & 97.03 & \textbf{97.36} & 98.54 & 97.98 & \textbf{98.65} & 98.54 \\
    la\_perseus & la\_proiel & 25,157 & 0.49 & 99.89 & 82.04 & 87.64 & \textbf{88.67} & 87.95 & 80.99 & 87.44 & \textbf{88.76} & 87.37 \\
    la\_proiel & la\_ittb & 174,977 & 0.20 & 99.98 & \textbf{95.30} & 95.15 & 94.83 & 95.10 & \textbf{96.65} & 94.54 & 96.03 & 95.47 \\
    lt\_hse & lv\_lvtb & 4,511 & 0.05 & 99.83 & 72.44 & 77.53 & \textbf{78.12} & 77.94 & 72.96 & 77.25 & \textbf{78.11} & 77.68 \\
    lv\_lvtb & hr\_set & 129,982 & 0.02 & 99.81 & \textbf{95.13} & 94.58 & 94.34 & 94.54 & 93.78 & 93.41 & 92.60 & \textbf{93.85} \\
    mr\_ufal & hi\_hdtb & 3,427 & 0.16 & 100.00 & 76.63 & 77.98 & 76.28 & \textbf{78.61} & 70.12 & 72.47 & \textbf{74.12} & 72.00 \\
    nl\_alpino & nl\_lassysmall & 178,169 & 0.23 & 93.27 & 95.88 & 95.49 & \textbf{96.18} & 95.98 & 95.60 & 94.57 & \textbf{95.61} & 95.43 \\
    nl\_lassysmall & nl\_alpino & 84,612 & 0.41 & 93.27 & 93.67 & 95.15 & \textbf{96.04} & 95.60 & 93.34 & 94.24 & \textbf{95.61} & 94.99 \\
    no\_bokmaal & no\_nynorsk & 264,958 & 0.24 & 96.34 & 97.24 & 97.41 & 97.25 & \textbf{97.65} & 98.23 & 97.86 & \textbf{98.31} & 98.01 \\
    no\_nynorsk & no\_bokmaal & 255,088 & 0.25 & 96.34 & 96.68 & 97.10 & 97.03 & \textbf{97.31} & 96.59 & 97.50 & \textbf{98.02} & 97.66 \\
    no\_nynorsklia & no\_nynorsk & 11,959 & 0.65 & 99.18 & 92.16 & 95.44 & \textbf{95.77} & 95.11 & 92.93 & \textbf{98.09} & 97.64 & 97.72 \\
    pcm\_nsc & en\_ewt & 11,038 & 0.74 & 99.99 & 92.95 & 93.28 & \textbf{94.26} & 94.15 & 98.55 & \textbf{99.92} & 99.84 & 99.76 \\
    pl\_lfg & pl\_sz & 118,526 & 0.41 & 60.73 & 95.71 & 94.43 & \textbf{96.43} & 93.91 & \textbf{95.93} & 94.67 & 95.49 & 94.96 \\
    pl\_sz & pl\_lfg & 73,011 & 0.53 & 60.73 & 92.68 & 88.62 & \textbf{94.82} & 89.33 & 95.78 & 93.96 & \textbf{95.79} & 94.81 \\
    pt\_bosque & pt\_gsd & 188,265 & 0.48 & 87.08 & 96.38 & 88.48 & \textbf{96.84} & 92.15 & 97.43 & 86.31 & \textbf{97.84} & 90.18 \\
    pt\_gsd & pt\_bosque & 265,352 & 0.41 & 87.08 & 97.63 & 94.92 & \textbf{98.03} & 93.49 & 97.53 & 94.37 & \textbf{98.36} & 94.38 \\
    ro\_nonstandard & ro\_rrt & 164,375 & 0.24 & 97.63 & 95.19 & 95.83 & 95.55 & \textbf{96.14} & 94.49 & \textbf{96.06} & 95.71 & 96.02 \\
    ro\_rrt & ro\_nonstandard & 182,366 & 0.11 & 97.63 & 97.25 & 97.16 & 97.19 & \textbf{97.30} & 97.09 & 97.11 & 96.47 & \textbf{97.36} \\
    ru\_gsd & ru\_syntagrus & 84,013 & 0.55 & 96.08 & 93.72 & 91.02 & \textbf{94.67} & 91.26 & 95.69 & 91.96 & \textbf{96.90} & 92.36 \\
    ru\_pud & ru\_syntagrus & 16,233 & 0.74 & 98.69 & 88.38 & 87.34 & \textbf{93.93} & 87.87 & 86.92 & \textbf{94.09} & 93.46 & 93.51 \\
    ru\_syntagrus & ru\_gsd & 937,395 & 0.13 & 96.08 & 96.69 & 96.84 & \textbf{97.36} & 96.40 & 96.73 & 96.42 & \textbf{97.50} & 95.68 \\
    ru\_taiga & ru\_syntagrus & 18,173 & 0.66 & 98.05 & 82.67 & \textbf{92.42} & 92.38 & 92.23 & 83.24 & 92.07 & 92.47 & \textbf{92.98} \\
    sa\_ufal & hi\_hdtb & 1,634 & 0.10 & 99.98 & 69.59 & 68.81 & 69.60 & \textbf{70.68} & 52.58 & 62.89 & 64.95 & \textbf{66.49} \\
    sk\_snk & cs\_pdt & 93,740 & 0.22 & 99.22 & \textbf{94.69} & 92.39 & 93.30 & 93.51 & \textbf{95.13} & 91.31 & 93.31 & 92.61 \\
    sl\_ssj & hr\_set & 118,536 & 0.12 & 99.33 & 95.38 & 94.98 & 95.09 & \textbf{95.65} & 96.13 & 95.58 & 95.97 & \textbf{96.32} \\
    sl\_sst & sl\_ssj & 26,309 & 0.55 & 99.20 & 89.49 & 92.91 & 93.47 & \textbf{93.75} & 91.76 & 95.64 & 95.93 & \textbf{96.38} \\
    sme\_giella & no\_nynorsk & 23,877 & 0.02 & 99.89 & 91.20 & 90.44 & \textbf{91.94} & 91.85 & 87.31 & 86.19 & \textbf{88.69} & \textbf{88.69} \\
    sr\_set & hr\_set & 72,045 & 0.62 & 88.98 & 95.52 & 95.24 & \textbf{97.47} & 96.09 & 95.77 & 95.96 & \textbf{97.02} & 95.96 \\
    sv\_lines & sv\_talbanken & 67,016 & 0.31 & 90.90 & 94.33 & 94.98 & \textbf{95.30} & 95.14 & \textbf{95.29} & 94.73 & 95.12 & 94.57 \\
    sv\_pud & sv\_talbanken & 15,758 & 0.39 & 93.08 & 89.65 & 93.78 & \textbf{95.12} & 94.20 & 83.86 & 88.30 & \textbf{93.11} & 90.35 \\
    sv\_talbanken & sv\_lines & 82,088 & 0.25 & 90.90 & 96.28 & 96.62 & \textbf{96.97} & 96.81 & \textbf{96.67} & 95.93 & 95.34 & 95.41 \\
    tl\_trg & es\_gsd & 274 & 0.13 & 99.98 & 74.73 & 82.22 & 76.92 & \textbf{83.15} & 60.00 & 72.00 & \textbf{80.00} & 76.00 \\
    tr\_imst & tr\_pud & 50,925 & 0.13 & 93.10 & 92.59 & 90.83 & \textbf{93.79} & 90.79 & 92.94 & 92.57 & \textbf{94.22} & 92.20 \\
    tr\_pud & tr\_imst & 14,180 & 0.33 & 93.10 & 91.82 & 86.88 & \textbf{93.83} & 87.66 & 84.80 & 84.92 & \textbf{86.32} & 84.34 \\
    uk\_iu & ru\_syntagrus & 98,865 & 0.10 & 99.55 & \textbf{92.69} & 91.14 & 92.41 & 92.13 & \textbf{93.67} & 91.59 & 93.07 & 92.86 \\
    ur\_udtb & fa\_seraji & 114,786 & 0.16 & 100.00 & \textbf{91.69} & 91.30 & 91.64 & 91.40 & \textbf{96.20} & 95.51 & 95.68 & 95.93 \\
    vi\_vtb & en\_ewt & 37,637 & 0.02 & 99.87 & \textbf{89.82} & 88.84 & 89.02 & 89.39 & 99.18 & 99.83 & 99.83 & \textbf{99.90} \\
    yo\_ytb & es\_gsd & 2,238 & 0.06 & 99.99 & 87.46 & 89.49 & 85.04 & \textbf{91.00} & 94.00 & 95.20 & 93.60 & \textbf{96.00} \\
    yue\_hk & zh\_gsd & 5,641 & 0.42 & 99.93 & 86.32 & 87.42 & \textbf{89.77} & 88.27 & 92.97 & \textbf{98.97} & \textbf{98.97} & \textbf{98.97} \\
    zh\_cfl & zh\_gsd & 6,048 & 0.34 & 99.93 & 86.15 & 88.92 & 88.13 & \textbf{89.72} & 91.00 & 95.57 & \textbf{96.26} & 95.98 \\
    zh\_gsd & ja\_gsd & 102,731 & 0.15 & 99.99 & 89.62 & \textbf{91.40} & 90.87 & 91.24 & 98.46 & 99.05 & 98.97 & \textbf{99.09} \\
    \midrule
    Average &  & 121,049 & 0.31 & 95.42 & 92.04 & 91.43 & \textbf{92.75} & 91.85 & 91.10 & 91.02 & \textbf{92.55} & 91.41 \\
    \bottomrule
    \end{tabular}}
    \caption{Joint morphological tagging and lemmatization results for all datasets. First column is the dataset for which the results are reported, the second column is the `helper' dataset. \textsc{size}: size of training data for the target dataset in words. \textsc{wo}: \% word overlap with `helper' dataset. \textsc{svm}: F1 score of svm classifier on dataset prediction. \textsc{base}: single dataset baseline. \textsc{concat}: simple concatenation of datasets. \textsc{gold}: using gold dataset embeddings. \textsc{pred}: predicted dataset embeddings.}
    \label{tab:fullMorph}
\end{table*}

\section{PCA-analysis of gold versus predicted treebank embeddings}
\begin{figure*}
\centering
\includegraphics[height=.45\textheight]{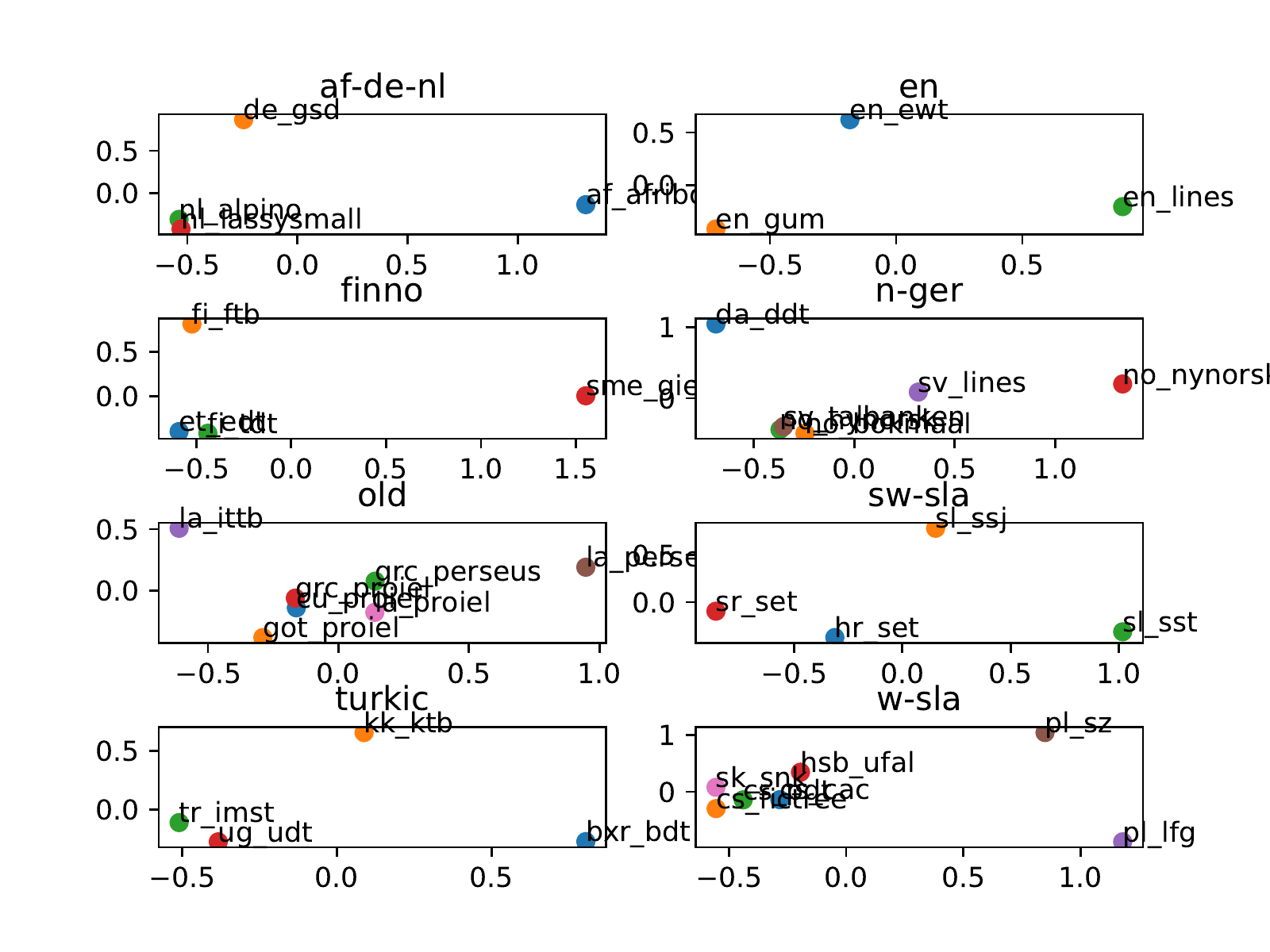}
\caption{PCA-projection of the gold dataset embeddings of the dataset groups.}
\label{fig:pcaGold}
\end{figure*}

\begin{figure*}
\centering
\includegraphics[height=.45\textheight]{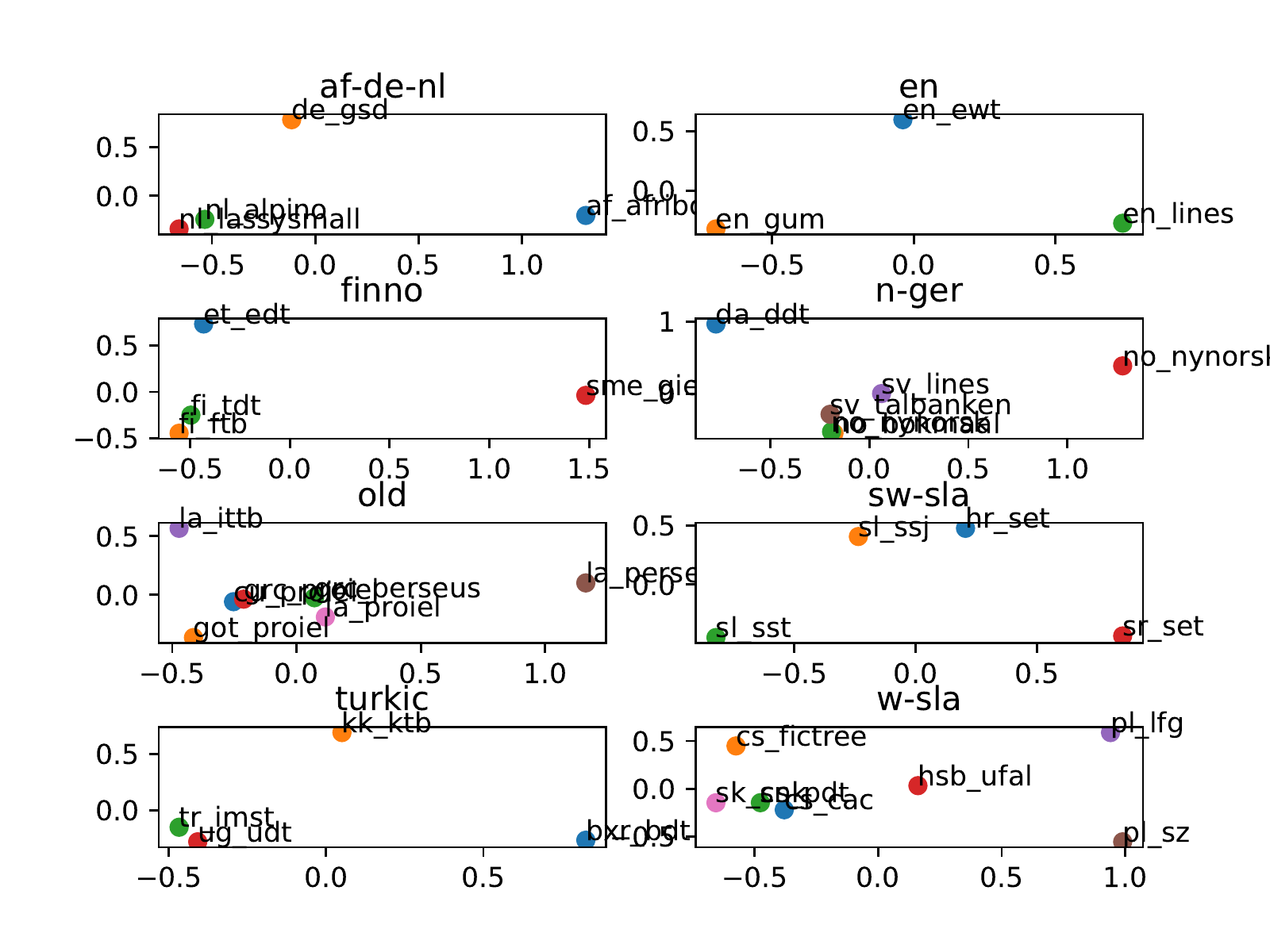}
\caption{PCA-projection of the predicted dataset embeddings of the dataset groups.}
\label{fig:pcaPred}
\end{figure*}

To gain deeper insights in what is represented in treebank embeddings, we
plotted the eight largest dataset groups of the UUParser setup into a PCA
space~\cite{pearson1901liii}. This is done with the default sklearn
settings~\cite{scikit-learn}. Results of the gold spaces are shown in
Figure~\ref{fig:pcaGold} and the predicted spaces are plotted in
Figure~\ref{fig:pcaPred}.  For some groups, there are some clear differences,
however for others the plots are highly similar. There seems to be no clear
trend in the amount of differences and the performance shifts in
Table~\ref{tab:fullDep}.


\end{document}